\documentclass[11pt,letterpaper]{article}
\usepackage{acl2012}
\usepackage{times}
\usepackage{latexsym}

\usepackage{mystuff-nlp}
\usepackage{url}
\usepackage{color}
\usepackage{float}
\makeatletter
\newcommand{\@BIBLABEL}{\@emptybiblabel}
\newcommand{\@emptybiblabel}[1]{}
\makeatother
\usepackage[hidelinks]{hyperref}
\usepackage{amsfonts,amssymb,amsmath,bm,bbm}
\usepackage{algorithm}
\usepackage{algpseudocode}
\usepackage{graphicx,subfigure}
\usepackage{booktabs}
\usepackage{multirow}

\usepackage{enumitem,array}

\newif\iffinal
\finaltrue

\newcolumntype{L}[1]{>{\raggedright\let\newline\\\arraybackslash\hspace{0pt}}m{#1}}

\newcommand{\mb}[1]{\mathbf{#1}}

\newcommand{\sysname}{\textsc{Social Attention}}

\newcommand{\example}[1]{`#1'}

\title{Overcoming Language Variation in Sentiment Analysis with Social Attention}

\iffinal
\author{Yi Yang \and Jacob Eisenstein\\
	    School of Interactive Computing\\
	    Georgia Institute of Technology\\
            Atlanta, GA 30308\\
	    {\tt \{yiyang+jacobe\}@gatech.edu}}
\else
\author{}
\fi

\date{}

\begin{document}
\maketitle
\begin{abstract}
Variation in language is ubiquitous, particularly in newer forms of writing such as social media. Fortunately, variation is not random; it is often linked to social properties of the author. In this paper, we show how to exploit social networks to make sentiment analysis more robust to social language variation. The key idea is \emph{linguistic homophily}: the tendency of socially linked individuals to use language in similar ways. We formalize this idea in a novel attention-based neural network architecture, in which attention is divided among several basis models, depending on the author's position in the social network. This has the effect of smoothing the classification function across the social network, and makes it possible to induce personalized classifiers even for authors for whom there is no labeled data or demographic metadata. This model significantly improves the accuracies of sentiment analysis on Twitter and on review data.


\end{abstract}

\section{Introduction}
\label{sec:intro}
Words can mean different things to different people. Fortunately, these differences are rarely idiosyncratic, but are often linked to social factors, such as age~\cite{rosenthal2011age}, gender~\cite{eckert2003language}, race~\cite{green2002african}, geography~\cite{trudgill1974linguistic}, and more ineffable characteristics such as political and cultural attitudes~\cite{fischer1958social,labov1963social}. In natural language processing (NLP), social media data has brought variation to the fore, spurring the development of new computational techniques for \emph{characterizing} variation in the lexicon~\cite{eisenstein2010latent}, orthography~\cite{eisenstein2015systematic}, and syntax~\cite{blodgett2016demographic}. However, aside from the focused task of spelling normalization~\cite{sproat2001normalization,aw2006phrase}, there have been few attempts to make NLP systems more robust to language variation across speakers or writers.

One exception is the work of \newcite{hovy2015demographic}, who shows that the accuracies of sentiment analysis and topic classification can be improved by the inclusion of coarse-grained author demographics such as age and gender. However, such demographic information is not directly available in most datasets, and it is not yet clear whether predicted age and gender offer any improvements. On the other end of the spectrum are attempts to create \emph{personalized} language technologies, as are often employed in information retrieval~\cite{shen2005implicit}, recommender systems~\cite{basilico2004unifying}, and language modeling~\cite{federico1996bayesian}. But personalization requires annotated data for each individual user---something that may be possible in interactive settings such as information retrieval, but is not typically feasible in natural language processing.

We propose a middle ground between group-level demographic characteristics and personalization, by exploiting social network structure. The sociological theory of \emph{homophily} asserts that individuals are usually similar to their friends~\cite{mcpherson2001birds}. This property has been demonstrated for language~\cite{bryden2013word} as well as for the demographic properties targeted by \newcite{hovy2015demographic}, which are more likely to be shared by friends than by random pairs of individuals~\cite{thelwall2009homophily}. Social network information is available in a wide range of contexts, from social media~\cite{huberman2008social} to political speech~\cite{thomas2006get} to historical texts~\cite{winterer2012america}. Thus, social network homophily has the potential to provide a more general way to account for linguistic variation in NLP. 

\begin{figure}[t!]
\centering
\includegraphics[scale=.34]{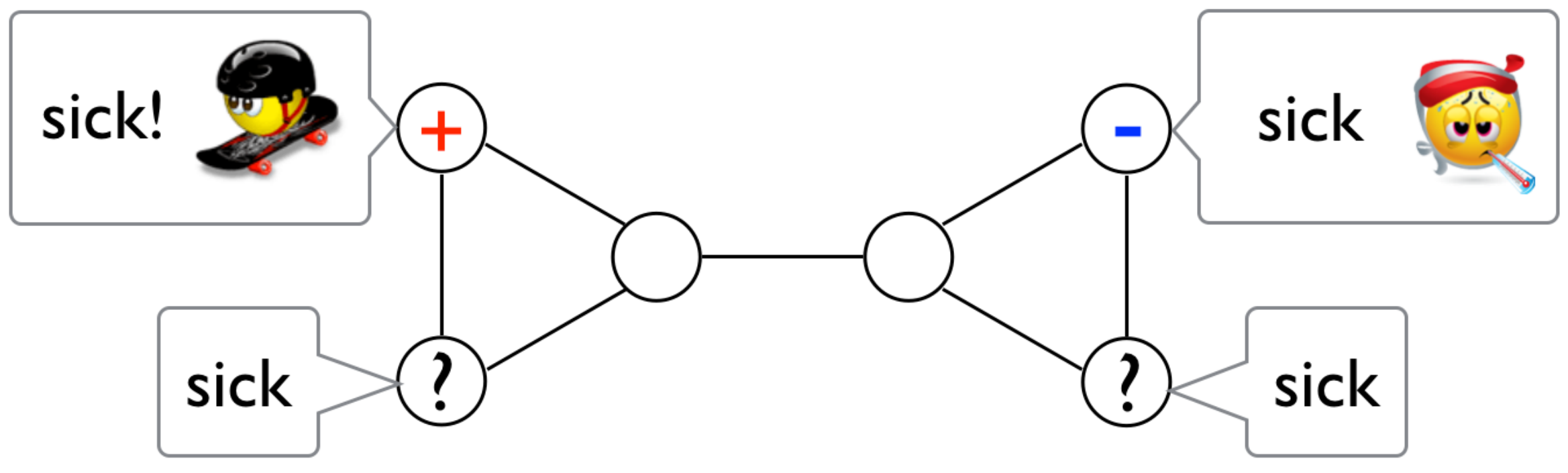}
\caption{Words such as \example{sick} can express opposite sentiment polarities depending on the author. We account for this variation by generalizing across the social network.}
\label{fig:sick}
\end{figure}

\autoref{fig:sick} gives a schematic of the motivation for our approach. The word \example{sick} typically has a negative sentiment, e.g., \example{I would like to believe he's sick rather than just mean and evil.}\footnote{Charles Rangel, describing Dick Cheney} However, in some communities the word can have a positive sentiment, e.g., the lyric \example{this sick beat}, recently trademarked by the musician Taylor Swift.\footnote{In the case of \example{sick}, speakers like Taylor Swift may employ either the positive and negative meanings, while speakers like Charles Rangel employ only the negative meaning. In other cases, communities may maintain completely distinct semantics for a word, such as the term \example{pants} in American and British English. Thanks to Christopher Potts for suggesting this distinction and this example.}
 Given labeled examples of \example{sick} in use by individuals in a social network, we assume that the word will have a similar sentiment meaning for their near neighbors---an assumption of \emph{linguistic homophily} that is the basis for this research. Note that this differs from the assumption of \emph{label homophily}, which entails that neighbors in the network will hold similar opinions, and will therefore produce similar document-level labels~\cite{tan2011user,hu2013exploiting}. Linguistic homophily is a more generalizable claim, which could in principle be applied to any language processing task where author network information is available.



To scale this basic intuition to datasets with tens of thousands of unique authors, we compress the social network into vector representations of each author node, using an embedding method for large scale networks~\cite{tang2015line}. Applying the algorithm to \autoref{fig:sick}, the authors within each triad would likely be closer to each other than to authors in the opposite triad. We then incorporate these embeddings into an attention-based neural network model, called \sysname, which employs multiple basis models to focus on different regions of the social network. 

We apply \sysname\ to Twitter sentiment classification, gathering social network metadata for Twitter users in the SemEval Twitter sentiment analysis tasks~\cite{nakov2013semeval}. We further adopt the system to Ciao product reviews~\cite{tang2012mtrust}, training author embeddings using trust relationships between reviewers. \sysname\ offers a 2-3\% improvement over related neural and ensemble architectures in which the social information is ablated. 
It also outperforms all prior published results on the SemEval Twitter test sets.



\section{Data}
\label{sec:data}

\begin{table} [t]
\centering
\small\addtolength{\tabcolsep}{-3pt}
\begin{tabular}{lrrrr}
    \toprule
    Dataset & \# Positive & \# Negative & \# Neutral & \# Tweet \\ \midrule
    Train 2013 & 3,230 & 1,265 & 4,109 & 8,604 \\
    Dev 2013  & 477 & 273 & 614 & 1,364 \\
    Test 2013 & 1,572 & 601 & 1,640 & 3,813 \\
    Test 2014 & 982 & 202 & 669 & 1,853 \\
    Test 2015 & 1,038 & 365 & 987 & 2,390 \\
    \bottomrule
\end{tabular}
\caption{Statistics of the SemEval Twitter sentiment datasets.}
\label{tab:data}
\end{table}

In the SemEval Twitter sentiment analysis tasks, the goal is to classify the sentiment of each message as positive, negative, or neutral. Following~\newcite{rosenthal2015semeval}, we train and tune our systems on the SemEval Twitter 2013 training and development datasets respectively, and evaluate on the 2013--2015 SemEval Twitter test sets. Statistics of these datasets are presented in Table~\ref{tab:data}. Our training and development datasets lack some of the original Twitter messages, which may have been deleted since the datasets were constructed. However, our test datasets contain all the tweets used in the SemEval evaluations, making our results comparable with prior work. 


We construct three author social networks based on the follow, mention, and retweet relations between the 7,438 authors in the training dataset, which we refer as  \textsc{Follower}, \textsc{Mention} and  \textsc{Retweet}.\footnote{We could not gather the authorship information of 10\% of the tweets in the training data, because the tweets or user accounts had been deleted by the time we crawled the social information.} Specifically, we use the Twitter API to crawl the friends of the SemEval users (individuals that they follow) and the most recent 3,200 tweets in their timelines.\footnote{The Twitter API returns a maximum of 3,200 tweets.} 
The mention and retweet links are then extracted from the tweet text and metadata. 
We treat all social networks as undirected graphs, where two users are socially connected if there exists at least one social relation between them.

\begin{figure*}
  \centering
  \includegraphics[width=0.9\linewidth]{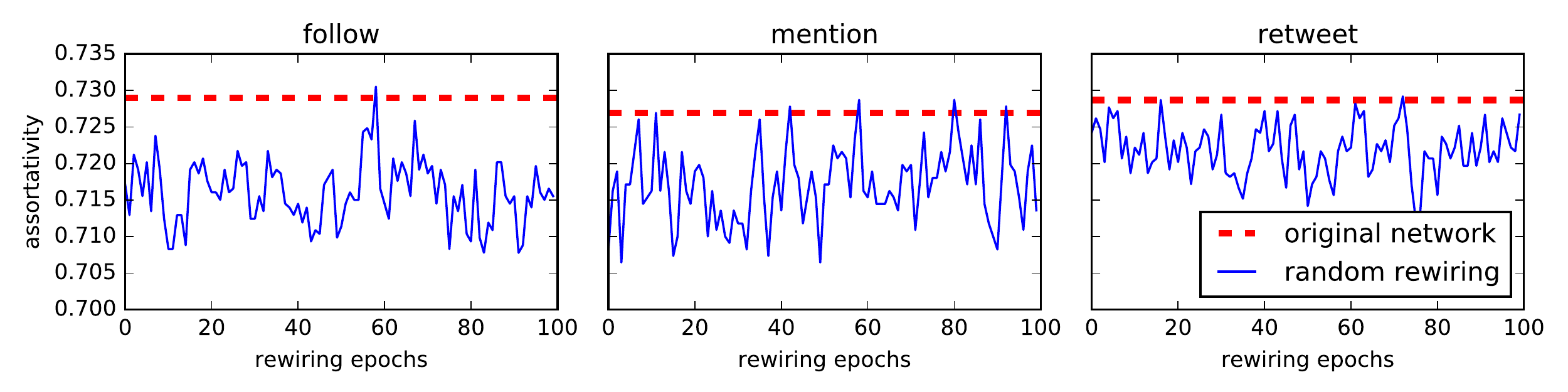}
  \caption{Assortativity of observed and randomized networks. Each rewiring epoch performs a number of rewiring operations equal to the total number of edges in the network. The randomly rewired networks almost always display lower assortativities than the original network, indicating that the accuracy of the lexicon-based sentiment analyzer is more assortative on the observed social network than one would expect by chance.}
  \label{fig:assortativity}
\end{figure*}

\section{Linguistic Homophily}
\label{sec:homophily}

The hypothesis of \emph{linguistic homophily} is that socially connected individuals tend to use language similarly, as compared to a randomly selected pair of individuals who are not socially connected. We now describe a pilot study that provides support for this hypothesis, focusing on the domain of sentiment analysis. The purpose of this study is to test whether errors in sentiment analysis are \emph{assortative} on the social networks defined in the previous section: that is, if two individuals $(i,j)$ are connected in the network, then a classifier error on $i$ suggests that errors on $j$ are more likely.

We test this idea using a simple lexicon-based classification approach, which we apply to the SemEval training data, focusing only on messages that are labeled as positive or negative (ignoring the neutral class), and excluding authors who contributed more than one message (a tiny minority). Using the social media sentiment lexicons defined by \newcite{tang2014building},\footnote{The lexicons include words that are assigned at least $0.99$ confidence by the method of \newcite{tang2014building}: 1,474 positive and 1,956 negative words in total.} we label a message as positive if it has at least as many positive words as negative words, and as negative otherwise.\footnote{Ties go to the positive class because it is more common.} The assortativity is the fraction of dyads for which the classifier makes two correct predictions or two incorrect predictions~\cite{newman2003structure}. This measures whether classification errors are clustered on the network.

We compare the observed assortativity against the assortativity in a network that has been randomly rewired.\footnote{Specifically, we use the \texttt{double\_edge\_swap} operation of the \texttt{networkx} package~\cite{hagberg2008exploring}. This operation preserves the degree of each node in the network.} 
Each rewiring epoch involves a number of random rewiring operations equal to the total number of edges in the network. 
(The edges are randomly selected, so a given edge may not be rewired in each epoch.) 
By counting the number of edges that occur in both the original and rewired networks, we observe that this process converges to a steady state after three or four epochs. 
As shown in~\autoref{fig:assortativity}, the original observed network displays more assortativity than the randomly rewired networks in nearly every case. 
Thus, the Twitter social networks display more linguistic homophily than we would expect due to chance alone. 

The differences in assortativity across network types are small, indicating that none of the networks are clearly best. The retweet network was the most difficult to rewire, with the greatest proportion of shared edges between the original and rewired networks. This may explain why the assortativities of the randomly rewired networks were closest to the observed network in this case.

\section{Model}
\label{sec:model}
 
In this section, we describe a neural network method that leverages social network information to improve text classification. Our approach is inspired by ensemble learning, where the system prediction is the weighted combination of the outputs of several basis models. We encourage each basis model to focus on a local region of the social network, so that classification on socially connected individuals employs similar model combinations.

Given a set of instances $\{ \mb{x}_i \}$ and authors $\{ a_i \}$, the goal of personalized probabilistic classification is to estimate a conditional label distribution $p(y \mid \mb{x}, a)$. For most authors, no labeled data is available, so it is impossible to estimate this distribution directly. We therefore make a smoothness assumption over a social network $G$: individuals who are socially proximate in $G$ should have similar classifiers. This idea is put into practice by modeling the conditional label distribution as a mixture over the predictions of $K$ basis classifiers,
\begin{small}
\begin{align}
p(y \mid \mb{x}, a) = \sum^K_{k=1} & \Pr(Z_a=k \mid a, G) \times p(y \mid \mb{x}, Z_a=k).
\label{eq:model-marginal}
\end{align}
\end{small}

The basis classifiers $p(y \mid \mb{x}, Z_a=k)$ can be arbitrary conditional distributions; we use convolutional neural networks, as described in \autoref{sec:model:cnn}. The component weighting distribution $\Pr(Z_a=k\mid a, G)$ is conditioned on the social network $G$, and functions as an attentional mechanism, described in \autoref{sec:model:attention}. The basic intuition is that for a pair of authors $a_i$ and $a_j$ who are nearby in the social network $G$, the prediction rules should behave similarly if the attentional distributions are similar, $p(z \mid a_i, G) \approx p(z \mid a_j, G).$ If we have labeled data only for $a_i$, some of the personalization from that data will be shared by $a_j$. The overall classification approach can be viewed as a mixture of experts~\cite{jacobs1991adaptive}, leveraging the social network as side information to choose the distribution over experts for each author.

\subsection{Social Attention Model}
\label{sec:model:attention}
The goal of the social attention model is to assign similar basis weights to authors who are nearby in the social network $G$. We operationalize social proximity by embedding each node's social network position into a vector representation. Specifically, we employ the \textsc{Line} method~\cite{tang2015line}, which estimates $D^{(v)}$ dimensional node embeddings $\mb{v}_a$ as parameters in a probabilistic model over edges in the social network. These embeddings are learned solely from the social network $G$, without leveraging any textual information. The attentional weights are then computed from the embeddings using a softmax layer,
\begin{equation}
\Pr(Z_a=k \mid a, G) = \frac{\exp\left(\mb{\phi}_k^{\top} \mb{v}_a + b_k\right)}{\sum_{k'}^K \exp\left(\mb{\phi}_{k'}^{\top} \mb{v}_a + b_{k'}\right)}.
\end{equation}

This embedding method uses only single-relational networks; in the evaluation, we will show results for Twitter networks built from networks of follow, mention, and retweet relations. In future work, we may consider combining all of these relation types into a unified multi-relational network. It is possible that embeddings in such a network could be estimated using techniques borrowed from multi-relational knowledge networks~\cite{bordes2013translating,wang2014knowledge}.

\subsection{Sentiment Classification with Convolutional Neural Networks}
\label{sec:model:cnn}
We next describe the basis models, $p(y \mid \mb{x}, Z = k)$. 
Because our target task is classification on microtext documents, we model this distribution using convolutional neural networks (CNNs; Lecun et al., 1989)\nocite{lecun1989backpropagation}, which have been proven to perform well on sentence classification tasks~\cite{kalchbrenner2014convolutional,kim2014convolutional}.
CNNs apply layers of convolving filters to n-grams, thereby generating a vector of dense local features. CNNs improve upon traditional bag-of-words models because of their ability to capture word ordering information.

Let $\mb{x} = [\mb{h}_1, \mb{h}_2, \cdots, \mb{h}_n]$ be the input sentence, where $\mb{h}_i$ is the $D^{(w)}$ dimensional word vector corresponding to the $i$-th word in the sentence. We use one convolutional layer and one max pooling layer to generate the sentence representation of $\mb{x}$.  The convolutional layer involves filters that are applied to bigrams to produce feature maps. Formally, given the bigram word vectors $\mb{h}_i, \mb{h}_{i+1}$, the features generated by $m$ filters can be computed by
\begin{equation}
\mb{c}_i = \tanh( \mb{W}_L \mb{h}_i + \mb{W}_R \mb{h}_{i+1} + \mb{b}),
\end{equation}
where $\mb{c}_i$ is an $m$ dimensional vector, $\mb{W}_L$ and $\mb{W}_R$ are $m \times D^{(w)}$ projection matrices, and $\mb{b}$ is the bias vector. The $m$ dimensional vector representation of the sentence is given by the pooling operation
\begin{equation}
\mb{s} = \max_{i\in 1, \cdots, n-1} \mb{c}_i .
\end{equation}


To obtain the conditional label probability, we utilize a multiclass logistic regression model,
\begin{equation}
\Pr(Y=t \mid \mb{x}, Z=k) = \frac{\exp(\bm{\beta}_t^\top \mb{s}_k + \beta_t)}{\sum_{t'=1}^T \exp(\bm{\beta}_{t'}^\top \mb{s}_k + \beta_{t'})},
\end{equation}
where $\bm{\beta}_t$ is an $m$ dimensional weight vector, $\beta_t$ is the corresponding bias term, and $\mb{s}_k$ is the $m$ dimensional sentence representation produced by the $k$-th basis model.

\subsection{Training}
We fix the pretrained author and word embeddings during training our social attention model. Let $\Theta$ denote the parameters that need to be learned, which include $\{\mb{W}_L, \mb{W}_R, \mb{b}, \{ \bm{\beta}_t, \beta_t \}_{t=1}^T \}$ for every basis CNN model, and the attentional weights $\{ \mb{\phi}_k, b_k \}_{k=1}^K$. We minimize the following logistic loss objective for each training instance:
\begin{equation}
\ell(\Theta) = - \sum_{t=1}^T \mb{1}[Y^*=t] \log \Pr(Y=t \mid \mb{x}, a),
\label{eq:loss}
\end{equation}
where $Y^*$ is the ground truth class for $\mb{x}$, and $\mb{1}[\cdot]$ represents an indicator function. We train the models for between $10$ and $15$ epochs using the Adam optimizer~\cite{kingma2014adam}, with early stopping on the development set.

\subsection{Initialization}
One potential problem is that after initialization, a small number of basis models may claim most of the mixture weights for all the users, while other basis models are inactive. This can occur because some basis models may be initialized with parameters that are globally superior. As a result, the ``dead'' basis models will receive near-zero gradient updates, and therefore can never improve. The true model capacity can thereby be substantially lower than the $K$ assigned experts.

Ideally, dead basis models will be avoided because each basis model should focus on a unique region of the social network. To ensure that this happens, we pretrain the basis models using an instance weighting approach from the domain adaptation literature~\cite{jiang2007instance}. For each basis model $k$, each author $a$ has an instance weight $\alpha_{a,k}$. These instance weights are based on the author's social network node embedding, so that socially proximate authors will have high weights for the same basis models. This is ensured by endowing each basis model with a random vector $\bm{\gamma}_k \sim N(\mb{0}, \sigma^2 \mathbbm{I})$, and setting the instance weights as,
\begin{equation}
\alpha_{a,k} = \text{sigmoid} (\bm{\gamma}_k^\top \mb{v}_a).
\end{equation}

The simple design results in similar instance weights for socially proximate authors. During pre-training, we train the $k$-th basis model by optimizing the following loss function for every instance:
\begin{small}\begin{equation}
\ell_k = - \alpha_{a,k} \sum_{t=1}^{T} \mb{1}[Y^*=t] \log \Pr(Y=t \mid \mb{x}, Z_a=k).
\end{equation}
\end{small}
The pretrained basis models are then assembled together and jointly trained using~\autoref{eq:loss}.

\section{Experiments}
Our main evaluation focuses on the 2013--2015 Sem\-Eval Twitter sentiment analysis tasks. The datasets have been described in~\autoref{sec:data}. We train and tune our systems on the Train 2013 and Dev 2013 datasets respectively, and evaluate on the Test 2013--2015 sets. In addition, we evaluate on another dataset based on Ciao product reviews~\cite{tang2012mtrust}.

\subsection{Social Network Expansion}
We utilize Twitter's follower, mention, and retweet social networks to train user embeddings. By querying the Twitter API in April 2015, we were able to identify 15,221 authors for the tweets in the SemEval datasets described above. We induce social networks for these individuals by crawling their friend links and timelines, as described in \autoref{sec:data}. Unfortunately, these networks are relatively sparse, with a large amount of isolated author nodes.
To improve the quality of the author embeddings, we expand the set of author nodes by adding nodes that do the most to densify the author networks: for the follower network, we add additional individuals that are followed by at least a hundred authors in the original set; for the mention and retweet networks, we add all users that have been mentioned or retweeted by at least twenty authors in the original set. The statistics of the resulting networks are presented in~\autoref{tab:networks}.

\begin{table} [t]
\centering
\begin{tabular}{lrr}
    \toprule
    Network & \# Author & \# Relation  \\ \midrule
    \textsc{Follower+} & 18,281 & 1,287,260  \\
    \textsc{Mention+} & 25,007 & 1,403,369 \\
    \textsc{Retweet+} & 35,376 & 2,194,319 \\
    \bottomrule
\end{tabular}
\caption{Statistics of the author social networks used for training author embeddings.}
\label{tab:networks}
\end{table}

\subsection{Experimental Settings}
We employ the pretrained word embeddings used by \newcite{astudillo2015learning}, which are trained with a corpus of 52 million tweets, and have been shown to perform very well on this task. The embeddings are learned using the structured skip-gram model~\cite{ling2015two}, and the embedding dimension is set at 600, following \newcite{astudillo2015learning}. We report the same evaluation metric as the SemEval challenge: the average F1 score of positive and negative classes.\footnote{Regarding the neutral class: systems are penalized with false positives when neutral tweets are incorrectly classified as positive or negative, and with false negatives when positive or negative tweets are incorrectly classified as neutral. This follows the evaluation procedure of the SemEval challenge.}

\begin{table*}[ht!]
\centering
\begin{tabular}{lllll}
    \toprule
    System & Test 2013 & Test 2014 & Test 2015 & Average \\ \midrule
    \multicolumn{4}{l}{\it Our implementations} \\
    CNN & 69.31 & 72.73 & 63.24 & 68.43 \\
    Mixture of experts & 68.97 & 72.07 & 64.28* & 68.44 \\
    Random attention & 69.48 & 71.56 & 64.37* & 68.47 \\
    Concatenation & 69.80 & 71.96 & 63.80 & 68.52 \\
    \sysname & 71.91* & \textbf{75.07}* & \textbf{66.75}* & \textbf{71.24} \\[4pt] 
    \multicolumn{4}{l}{\it Reported results} \\
    \textsc{Nlse} & 72.09 & 73.64 & 65.21 & 70.31 \\
    \textsc{Webis} & 68.49 & 70.86 & 64.84 & 68.06 \\
    \textsc{unitn} & \textbf{72.79} & 73.60 & 64.59 & 70.33 \\
    \textsc{lsislif} & 71.34 & 71.54 & 64.27 & 69.05 \\
    \bottomrule
\end{tabular}
\caption{Average F1 score on the SemEval test sets. The best results are in {\bf bold}. Results are marked with * if they are significantly better than CNN at $p<0.05$.}
\label{tab:results}
\end{table*}

\begin{table}
\centering
\begin{tabular}{lllll}
    \toprule
  & \multicolumn{3}{c}{SemEval Test} \\
    Network & 2013 & 2014 & 2015 & Average \\ \midrule
    \textsc{Follower+} & 71.49 & 74.17 & 66.00 & 70.55 \\
    \textsc{Mention+}  & 71.72 & 74.14 & 66.27 & 70.71 \\
    \textsc{Retweet+}  & \textbf{71.91} & \textbf{75.07} & \textbf{66.75} & \textbf{71.24}  \\
    \bottomrule
\end{tabular}
\caption{Comparison of different social networks with \sysname. The best results are in {\bf bold}.}
\label{tab:user}
\end{table}

\paragraph{Competitive systems}
We consider five competitive Twitter sentiment classification methods. \emph{Convolutional neural network} (CNN) has been described in~\autoref{sec:model:cnn}, and is the basis model of \sysname. \emph{Mixture of experts} employs the same CNN model as an expert, but the mixture densities solely depend on the input values. We adopt the summation of the pretrained word embeddings as the sentence-level input to learn the gating function.\footnote{The summation of the pretrained word embeddings works better than the average of the word embeddings.} The model architecture of \emph{random attention} is nearly identical to \sysname: the only distinction is that we replace the pretrained author embeddings with random embedding vectors, drawing uniformly from the interval $(-0.25, 0.25)$. \emph{Concatenation} concatenates the author embedding with the sentence representation obtained from CNN, and then feeds the new representation to a softmax classifier. Finally, we include \sysname, the attention-based neural network method described in~\autoref{sec:model}. 

We also compare against the three top-performing systems in the SemEval 2015 Twitter sentiment analysis challenge~\cite{rosenthal2015semeval}: \textsc{Webis}~\cite{hagen2015webis}, \textsc{unitn}~\cite{severynunitn}, and \textsc{lsislif}~\cite{hamdan2015lsislif}. \textsc{unitn} achieves the best average F1 score on Test 2013--2015 sets among all the submitted systems. Finally, we republish results of \textsc{Nlse}~\cite{astudillo2015learning}, a non-linear subspace embedding model.

\paragraph{Parameter tuning}
We tune all the hyperparameters on the SemEval 2013 development set. We choose the number of bigram filters for the CNN models from \{50, 100, 150\}. The size of author embeddings is selected from \{50, 100\}.  For \emph{mixture of experts}, \emph{random attention} and \emph{\sysname}, we compare a range of numbers of basis models, \{3, 5, 10, 15\}. We found that a relatively small number of basis models are usually sufficient to achieve good performance.
The number of pretraining epochs is selected from \{1, 2, 3\}. During joint training, we check the performance on the development set after each epoch to perform early stopping.

 \begin{table*} [t]
\centering
\begin{tabular}{lll}
    \toprule
    Basis model & More positive & More negative \\ \midrule
    1 & banging loss fever broken \underline{fucking} & {\bf dear like} god {\bf yeah wow} \\ 
    2 & chilling cold ill sick suck & satisfy trust wealth strong {\bf lmao} \\
    3 & \underline{ass} \underline{damn} \underline{piss} \underline{bitch} \underline{shit} & talent honestly voting win clever \\ 
    4 & insane bawling fever weird cry & {\bf lmao} super {\bf lol haha hahaha} \\
    5 & ruin silly bad boring dreadful & {\it lovatics} wish {\it beliebers arianators kendall} \\ 
    \bottomrule
\end{tabular}
\caption{Top 5 more positive/negative words for the basis models in the SemEval training data. {\bf Bolded} entries correspond to words that are often used ironically,  by top authors related to basis model 1 and 4. \underline{Underlined} entries are swear words, which are sometimes used positively by top users corresponding to basis model 3. {\it Italic} entries refer to celebrities and their fans, which usually appear in negative tweets by top authors for basis model 5.}
\label{tab:words}
\end{table*}

\begin{table*} [t]
\centering
\small\addtolength{\tabcolsep}{-2pt}
\begin{tabular}{l l L{12cm}}
    \toprule
    Word & Sentiment & Example \\ \midrule
    sick & positive & Watch ESPN tonight to see me burning @user for a sick goal on the top ten. \#realbackyardFIFA \\ [1.5pt]
    bitch & positive & @user bitch u shoulda came with me Saturday sooooo much fun. Met Romeo santos lmao na i met his look a like \\ [1.5pt]
    shit & positive & @user well shit! I hope your back for the morning show. I need you on my drive to Cupertino on Monday! Have fun! \\[3pt]
    dear & negative & Dear Spurs, You are out of COC, not in Champions League and come May wont be in top 4. Why do you even exist? \\  [1.5pt]
    wow & negative & Wow. Tiger fires a 63 but not good enough.  Nick Watney shoots a 59 if he birdies the 18th?!? \#sick \\ [1.5pt]
    lol & negative & Lol super awkward if its hella foggy at Rim tomorrow and the games suppose to be on tv lol  Uhhhh.. Where's the ball? Lol \\ [1.5pt]
    \bottomrule
\end{tabular}
\caption{Tweet examples that contain sentiment words conveying specific sentiment meanings that differ from their common senses in the SemEval training data. The sentiment labels are adopted from the SemEval annotations.}
\label{tab:tweets}
\end{table*}

\subsection{Results}
\autoref{tab:results} summarizes the main empirical findings, where we report results obtained from author embeddings trained on \textsc{Retweet+} network for \sysname. The results of different social networks for \sysname\ are shown in~\autoref{tab:user}. The best hyperparameters are: $100$ bigram filters; $100$-dimensional author embeddings; $K=5$ basis models; $1$ pre-training epoch. To establish the statistical significance of the results, we obtain 100 bootstrap samples for each test set, and compute the F1 score on each sample for each algorithm.  A two-tail paired t-test is then applied to determine if the F1 scores of two algorithms are significantly different, $p < 0.05$.

Mixture of experts, random attention, and CNN all achieve similar average F1 scores on the SemEval Twitter 2013--2015 test sets. Note that random attention can benefit from some of the personalized information encoded in the random author embeddings, as Twitter messages posted by the same author share the same attentional weights. However, it barely improves the results, because the majority of authors contribute a single message in the SemEval datasets. With the incorporation of author social network information, concatenation slightly improves the classification performance. Finally, \sysname\ gives much better results than concatenation, as it is able to model the interactions between text representations and author representations. It significantly outperforms CNN on all the SemEval test sets, yielding 2.8\% improvement on average F1 score. \sysname\ also performs substantially better than the top-performing SemEval systems and \textsc{Nlse}, especially on the 2014 and 2015 test sets.

We now turn to a comparison of the social networks. As shown in~\autoref{tab:user}, the \textsc{Retweet+} network is the most effective, although the differences are small: \sysname\ outperforms prior work regardless of which network is selected. Twitter's ``following'' relation is a relatively low-cost form of social engagement, and it is less public than retweeting or mentioning another user. Thus it is unsurprising that the follower network is least useful for socially-informed personalization. 
The \textsc{Retweet+} network has denser social connections than \textsc{Mention+}, which could lead to better author embeddings.

\subsection{Analysis}

We now investigate whether language variation in sentiment meaning has been captured by different basis models. We focus on the same sentiment words~\cite{tang2014building} that we used to test linguistic homophily in our analysis. We are interested to discover sentiment words that are used with the opposite sentiment meanings by some authors. To measure the level of model-specificity for each word $w$, we compute the difference between the model-specific probabilities $p(y \mid X=w, Z=k)$ and the average probabilities of all basis models $\frac{1}{K} \sum_{k=1}^K p(y \mid X=w, Z=k)$ for positive and negative classes. The five words in the negative and positive lexicons with the highest scores for each model are presented in~\autoref{tab:words}. 

As shown in~\autoref{tab:words}, Twitter users corresponding to basis models 1 and 4 often use some words ironically in their tweets. Basis model 3 tends to assign positive sentiment polarity to swear words, and Twitter users related to basis model 5 seem to be less fond of fans of certain celebrities. Finally, basis model 2 identifies Twitter users that we have described in the introduction---they often adopt general negative words like \example{ill}, \example{sick}, and \example{suck} positively. Examples containing some of these words are shown in~\autoref{tab:tweets}.

\subsection{Sentiment Analysis of Product Reviews}
The labeled datasets for Twitter sentiment analysis are relatively small; to evaluate our method on a larger dataset, we utilize a product review dataset by~\newcite{tang2012mtrust}.  The dataset consists of 257,682 reviews written by 10,569 users crawled from a popular product review sites, Ciao.\footnote{\url{http://www.ciao.co.uk}} The rating information in discrete five-star range is available for the reviews, which is treated as the ground truth label information for the reviews. Moreover, the users of this site can mark explicit ``trust'' relationships with each other, creating a social network. 

To select examples from this dataset, we first removed reviews that were marked by readers as ``not useful.'' We treated reviews with more than three stars as positive, and less than three stars as negative; reviews with exactly three stars were removed. We then sampled 100,000 reviews from this set, and split them randomly into training (70\%), development (10\%) and test sets (20\%). The statistics of the resulting datasets are presented in~\autoref{tab:ciao}. We utilize 145,828 trust relations between 18,999 Ciao users to train the author embeddings. We consider the 10,000 most frequent words in the datasets, and assign them pretrained word2vec embeddings.\footnote{\url{https://code.google.com/archive/p/word2vec}} As shown in~\autoref{tab:ciao}, the datasets have highly skewed class distributions. Thus, we use the average F1 score of positive and negative classes as the evaluation metic.

\begin{table} [t]
\centering
\small\addtolength{\tabcolsep}{-3pt}
\begin{tabular}{lrrrr}
    \toprule
    Dataset & \# Author & \# Positive & \# Negative & \# Review\\ \midrule
    Train Ciao & 8,545 & 63,047 & 6,953 & 70,000 \\
    Dev Ciao  & 4,087 & 9,052 & 948 & 10,000 \\
    Test Ciao & 5,740 & 17,978 & 2,022 & 20,000 \\ [3pt]
    Total & 9,267 & 90,077 & 9,923 & 100,000 \\
    \bottomrule
\end{tabular}
\caption{Statistics of the Ciao product review datasets.}
\label{tab:ciao}
\end{table}

\begin{table} [t]
\centering
\begin{tabular}{ll}
    \toprule
    System & Test Ciao \\ \midrule
    CNN &  78.43 \\
    Mixture of experts  &  78.37 \\
    Random attention & 79.43* \\
    Concatenation & 77.99 \\
    \sysname & \textbf{80.19}** \\
    \bottomrule
\end{tabular}
\caption{Average F1 score on the Ciao test set. The best results are in {\bf bold}. Results are marked with * and ** if they are significantly better than CNN and random attention respectively, at $p<0.05$.}
\label{tab:ciao-results}
\end{table}

The evaluation results are presented in~\autoref{tab:ciao-results}. The best hyperparameters are generally the same as those for Twitter sentiment analysis, except that the optimal number of basis models is $10$, and the optimal number of pretraining epochs is $2$.  
Mixture of experts and concatenation obtain slightly worse F1 scores than the baseline CNN system, but random attention performs significantly better. In contrast to the SemEval datasets, individual users often contribute multiple reviews in the Ciao datasets (the average number of reviews from an author is 10.8; \autoref{tab:ciao}). As an author tends to express similar opinions toward related products, random attention is able to leverage the personalized information to improve sentiment analysis. Prior work has investigated the direction, obtaining positive results using speaker adaptation techniques~\cite{al2015model}. Finally, by exploiting the social network of trust relations, \sysname\ obtains further improvements, outperforming random attention by a small but significant margin.

\section{Related Work}

\paragraph{Domain adaptation and personalization}
Domain adaptation is a classic approach to handling the variation inherent in social media data~\cite{eisenstein2013bad}. Early approaches to supervised domain adaptation focused on adapting the classifier weights across domains, using enhanced feature spaces~\cite{daume2007frustratingly} or Bayesian priors~\cite{chelba2006adaptation,finkel2009hierarchical}. Recent work focuses on unsupervised domain adaptation, which typically works by transforming the input feature space so as to overcome domain differences
~\cite{blitzer2006domain}. 
However, in many cases, the data has no natural partitioning into domains. In preliminary work, we constructed social network domains by running community detection algorithms on the author social network~\cite{fortunato2010community}. However, these algorithms proved to be unstable on the sparse networks obtained from social media datasets, and offered minimal performance improvements. In this paper, we convert social network positions into node embeddings, and use an attentional component to smooth the classification rule across the embedding space. 


Personalization has been an active research topic in areas such as speech recognition and information retrieval. Standard techniques for these tasks include linear transformation of model parameters~\cite{leggetter1995maximum} and collaborative filtering~\cite{breese1998empirical}. These methods have recently been adapted to personalized sentiment analysis~\cite{tang2015learning,al2015model}. Supervised personalization typically requires labeled training examples for every individual user. In contrast, by leveraging the social network structure, we can obtain personalization even when labeled data is unavailable for many authors.

\paragraph{Sentiment analysis with social relations}
Previous work on incorporating social relations into sentiment classification has relied on the label consistency assumption, where the existence of social connections between users is taken as a clue that the sentiment polarities of the users' messages should be similar. \newcite{speriosu2011twitter} construct a heterogeneous network with tweets, users, and n-grams as nodes. Each node is then associated with a sentiment label distribution, and these label distributions are smoothed by label propagation over the graph.
Similar approaches are explored by \newcite{hu2013exploiting}, who employ the graph Laplacian as a source of regularization, and by \newcite{tan2011user} who take a factor graph approach.
A related idea is to label the sentiment of individuals in a social network towards each other: \newcite{west2014exploiting} exploit the sociological theory of structural balance to improve the accuracy of dyadic sentiment labels in this setting. All of these efforts are based on the intuition that individual predictions $p(y)$ should be smooth across the network. In contrast, our work is based on the intuition that social neighbors use language similarly, so they should have a similar \emph{conditional} distribution $p(y \mid x)$. These intuitions are complementary: if both hold for a specific setting, then label consistency and linguistic consistency could in principle be combined to improve performance.

Social relations can also be applied to improve personalized sentiment analysis~\cite{song2015personalized,wu2016personalized}. 
\newcite{song2015personalized} present a latent factor model that alleviates the data sparsity problem by decomposing the messages into words that are represented by the weighted sentiment and topic units. Social relations are further incorporated into the model based on the intuition that linked individuals share similar interests with respect to the latent topics. \newcite{wu2016personalized} build a personalized sentiment classifier for each author; socially connected users are encouraged to have similar user-specific classifier components. As discussed above, the main challenge in personalized sentiment analysis is to obtain labeled data for each individual author. Both papers employ distant supervision, using emoticons to label additional instances. However, emoticons may be unavailable for some authors or even for entire genres, such as reviews. Furthermore, the pragmatic function of emoticons is complex, and in many cases emoticons do not refer to sentiment~\cite{walther2001impacts}. Our approach does not rely on distant supervision, and assumes only that the classification decision function should be smooth across the social network.
\section{Conclusion}
This paper presents a new method for learning to overcome language variation, leveraging the tendency of socially proximate individuals to use language similarly---the phenomenon of \emph{linguistic homophily}. By learning basis models that focus on different local regions of the social network, our method is able to capture subtle shifts in meaning across the network. 
Inspired by ensemble learning, we have formulated this model by employing a social attention mechanism: the final prediction is the weighted combination of the outputs of the basis models, and each author has a unique weighting, depending on their position in the social network. 
Our model achieves significant improvements over standard convolutional networks, and ablation analyses show that social network information is the critical ingredient. In other work, language variation has been shown to pose problems for the entire NLP stack, from part-of-speech tagging to information extraction. A key question for future research is whether we can learn a socially-infused ensemble that is useful across multiple tasks.

\iffinal
\section{Acknowledgments}

We thank Duen Horng ``Polo'' Chau for discussions about community detection and Ramon Astudillo for sharing data and helping us to reproduce the NLSE results. This research was supported by the National Science Foundation under award RI-1452443, by the National Institutes of Health under award number R01GM112697-01, and by the Air Force Office of Scientific Research. The content is solely the responsibility of the authors and does not necessarily represent the official views of these sponsors.

\else
\fi

\bibliographystyle{acl2012}
\bibliography{cite-strings,cites,cite-definitions}

\end{document}